\documentclass[11pt]{article}
\usepackage[margin=1in]{geometry}
    \PassOptionsToPackage{numbers, compress}{natbib}
\usepackage{helvet}  %
\usepackage{courier}  %
\usepackage{hyperref}
\usepackage{url}
\usepackage{graphicx} %
\usepackage{caption} %
\usepackage{amsmath,amsfonts,bm}

\def\eqref#1{equation~\ref{#1}}

\def\1{\bm{1}}

\def\vf{{\bm{f}}}

\def\vp{{\bm{p}}}

\DeclareMathAlphabet{\mathsfit}{\encodingdefault}{\sfdefault}{m}{sl}
\SetMathAlphabet{\mathsfit}{bold}{\encodingdefault}{\sfdefault}{bx}{n}

\usepackage[capitalize]{cleveref}
\usepackage{newfloat}
\usepackage{listings}
\usepackage{xurl}
\usepackage[ruled,vlined,linesnumbered]{algorithm2e}
\crefname{algocf}{Algorithm}{Algorithms}
\Crefname{algocf}{Algorithm}{Algorithms}
\usepackage{amssymb}
\usepackage{booktabs}
\usepackage{colortbl}
\usepackage{multirow}
\usepackage{multicol}
\usepackage{makecell}
\usepackage{arydshln}
\usepackage{enumitem}
\usepackage{xspace}
\usepackage{natbib}

\newcommand{\blankbox}{\vcenter{\hbox{\includegraphics[scale=0.39]{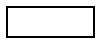}}}}
\newcommand{\slashbox}{\vcenter{\hbox{\includegraphics[scale=0.39]{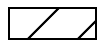}}}}

\newcommand{\german}{\texttt{German}\xspace}
\newcommand{\bail}{\texttt{Bail}\xspace}
\newcommand{\credit}{\texttt{Credit}\xspace}
\newcommand{\pokecz}{\texttt{Pokec-z}\xspace}
\newcommand{\pokecn}{\texttt{Pokec-n}\xspace}
\newcommand{\nba}{\texttt{NBA}\xspace}

\newcommand{\graph}{\ensuremath{\mathcal{G}}\xspace}

\newcommand{\feat}{\mathbf{X}\xspace}
\newcommand{\adj}{\mathbf{A}\xspace}
\newcommand{\sens}{\mathbf{S}\xspace}
\newcommand{\vecp}{\mathbf{P}\xspace}
\newcommand{\vecf}{\mathbf{F}\xspace}

\newcommand{\vecn}{\mathbf{N}\xspace}

\newcommand{\nlabel}{\mathbf{Y}\xspace}

\newcommand{\method}{\textsf{Fair\-GHDC}}

\newcommand{\ms}[2]{{#1\footnotesize{$\pm$#2}}}

\makeatletter
\newenvironment{fullitemize}
{
\vspace{-1pt}
\begin{itemize}[leftmargin=*]
\setlength{\itemsep}{5pt}
\setlength{\parsep}{-5pt}
\setlength{\parskip}{-3pt}
\setlength{\leftmargin}{-10pt}
}
{
\end{itemize}
\vspace{-1pt}
}
\makeatother

\DeclareCaptionStyle{ruled}{labelfont=normalfont,labelsep=colon,strut=off} %
\usepackage{xcolor}    
\definecolor{tablehead}{RGB}{121,80,242}    
\title{Mitigating Bias in Graph Hyperdimensional Computing}
\author{
\textbf{Yezi Liu}\textsuperscript{1} \quad
\textbf{William Youngwoo Chung}\textsuperscript{1} \quad
\textbf{Yang Ni}\textsuperscript{2} \\ 
  \textbf{Hanning Chen}\textsuperscript{1} \quad
  \textbf{Mohsen Imani}\textsuperscript{1} \\
  \textsuperscript{1}University of California, Irvine \quad 
  \textsuperscript{2}Purdue University Northwest \quad \\
\texttt{\{yezil3,chungwy1,hanningc,m.imani\}@uci.edu} \\
\texttt{yangni@purdue.edu}
}
\date{}  

\begin{document}

\maketitle

\begin{abstract}
Graph hyperdimensional computing (HDC) has emerged as a promising paradigm for cognitive tasks, emulating brain-like computation with high-dimensional vectors known as hypervectors. While HDC offers robustness and efficiency on graph-structured data, its fairness implications remain largely unexplored. In this paper, we study fairness in graph HDC, where biases in data representation and decision rules can lead to unequal treatment of different groups. We show how hypervector encoding and similarity-based classification can propagate or even amplify such biases, and we propose a fairness-aware training framework, {\method}, to mitigate them. {\method} introduces a bias correction term, derived from a gap-based demographic-parity regularizer, and converts it into a scalar fairness factor that scales the update of the class hypervector for the ground-truth label. This enables debiasing directly in the hypervector space without modifying the graph encoder or requiring backpropagation. Experimental results on six benchmark datasets demonstrate that {\method} substantially reduces demographic-parity and equal-opportunity gaps while maintaining accuracy comparable to standard GNNs and fairness-aware GNNs. At the same time, {\method} preserves the computational advantages of HDC, achieving up to about one order of magnitude (\(\approx 10\times\)) speedup in training time on GPU compared to GNN and fairness-aware GNN baselines.
\end{abstract}

\section{Introduction}
Graph Neural Networks (GNNs) have garnered considerable attention in recent years for their effectiveness in various tasks such as social network analysis~\citep{hamilton2017inductive,wu2020comprehensive}, recommendation systems~\citep{ying2018graph,battaglia2018relational,zhou2018graph}, drug discovery~\citep{duvenaud2015convolutional,bongini2021molecular,jiang2021could}, and epidemiology~\citep{liu2024review}. While GNNs have demonstrated success in capturing rich information from graph data, training them on real-world graphs, often consisting of large numbers of nodes and edges, can be computationally expensive and time-consuming. This challenge arises from the complex sparse matrix operations required during training. Additionally, the high-dimensional features commonly found in graph data exacerbate the issue. The demands for both storage and computational resources are even more pronounced in automated machine learning settings~\citep{cai2022multimodal}, such as neural architecture search and hyperparameter optimization, where GNNs need to be retrained multiple times.

However, training GNNs involves backpropagation over irregular, sparse graph structures, leading to high computational complexity. In contrast, brain-inspired Hyperdimensional Computing (HDC) offers a more efficient and lightweight alternative. Based on Sparse Distributed Memory~\citep{kanerva1988sparse}, HDC models human cognition using high-dimensional vectors. Applications of HDC, including language recognition~\citep{rahimi2016robust}, activity identification, speech recognition~\citep{imani2017voicehd}, and recommendation systems~\citep{guo2021hyperrec}, have shown performance comparable to traditional machine learning methods without the need for compute-intensive backpropagation. 

HDC models data using high-dimensional (HD) vectors, referred to as hypervectors (HVs)~\citep{kanerva2009hyperdimensional}, and simulates brain-like reasoning by applying arithmetic operations on these HVs. Several previous works have applied the HDC paradigm to graph-related tasks. For instance, in~\citep{kleyko2021vector}, edges are represented in HD space by binding the HVs of the source and destination nodes. In another approach,~\citep{ma2018holistic} employed holographic reduced representation to map nodes into HD space, though this method required a neural network during inference. More recently,~\citep{poduval2022graphd} leveraged the memorization capability of HDC to encode graph structures into HVs, enabling both information retrieval and graph reconstruction based on memorized information. Similarly,~\citep{nunes2022graphhd} encoded entire graphs into HVs to perform graph classification.

Despite advancements in graph HDC, these models often inherit or even amplify biases present in the training data, resulting in discriminatory predictions against groups defined by sensitive attributes such as race and gender~\citep{kipf2016semi,hamilton2017inductive,zhang2018graph,zhou2020graph,wu2020comprehensive}. This presents significant ethical challenges, particularly in critical applications like job applicant ranking~\citep{mehrabi2021survey}, loan fraud detection~\citep{xu2021towards}, and crime prediction~\citep{suresh2019framework}, where fairness is essential. To address these concerns, recent work~\citep{dai2021say,ling2023learning,kose2022fair} has focused on reducing bias in graph representation models, promoting fairness, and facilitating their responsible and reliable use in real-world scenarios.

In this study, we propose a fairness-aware graph hyperdimensional computing approach, {\method}, designed to accelerate training for node classification tasks in graph learning models. Fairness is achieved by dynamically updating the class hypervectors (HVs). Specifically, {\method} first computes the demographic parity difference among the batch of nodes, referred to as the bias correction. This bias correction is then applied to the class HVs associated with the predicted labels, ensuring fairness in training the class HVs. Our contribution can be summarized as follows:

\begin{figure*}[t]
\centering
\includegraphics[width=1.0\linewidth]{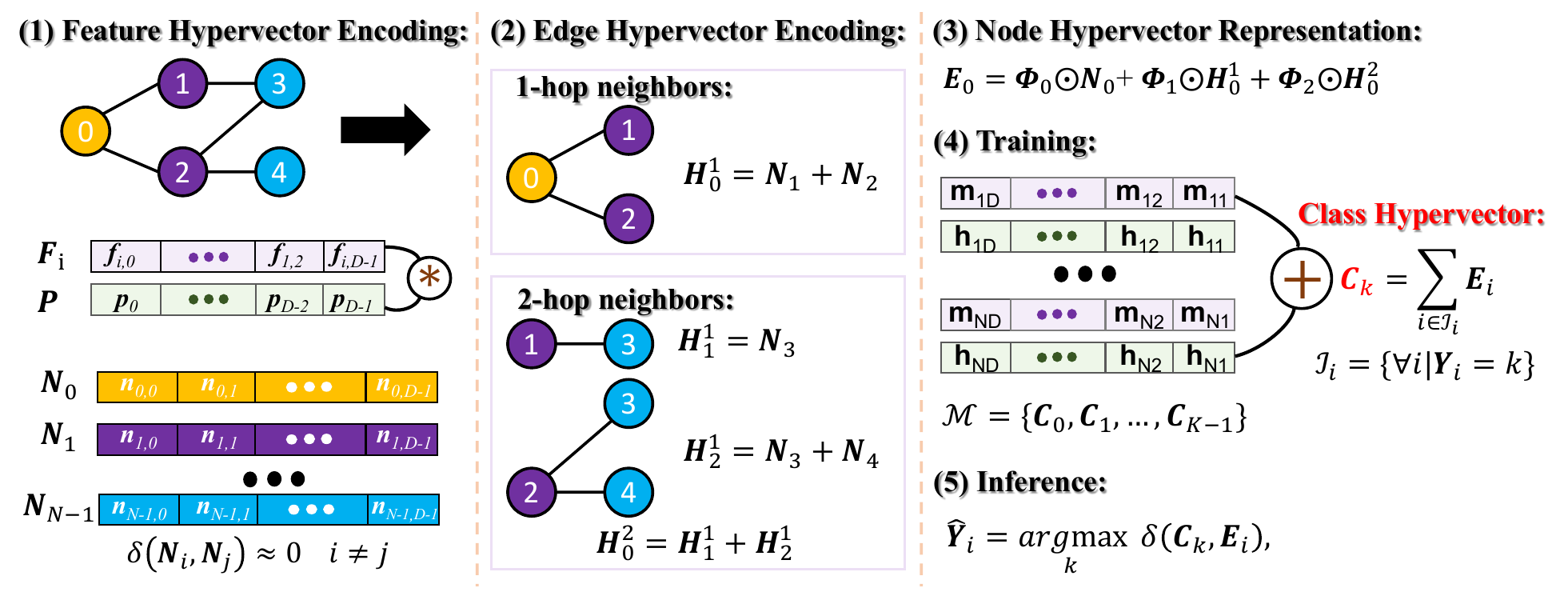}
\vspace{-2em}
\caption{The overall framework of graph HDC in {\method}, including four steps: (1) Feature Hypervector Encoding, (2) Edge Hypervector Encoding, (3) Node Hypervector Representation, (4) Training, and (5) Inference.} \label{fig:framework}
\vspace{-1em}
\end{figure*}
\begin{fullitemize}
    \item To the best of our knowledge, this is the first work to systematically study group fairness in graph hyperdimensional computing. We characterize how bias can arise in HDC-based graph classifiers and provide a framework and empirical analysis that lay the groundwork for future research on fair and trustworthy HDC-based graph learning.
    \item We introduce {\method}, a method that improves group fairness while maintaining competitive node classification accuracy, and is efficient and scalable, making it suitable for large-scale graphs.
    \item Our experimental results, conducted on six graph datasets used in fairness studies, demonstrate that {\method} often achieves superior fairness compared to fairness-aware GNNs, while also maintaining satisfactory node classification accuracy. Additionally, the runtime experiments underscore {\method}'s ability to significantly accelerate training on large-scale graphs.
\end{fullitemize}

\section{Preliminary}
In this section, we provide the necessary background on fairness in graph HDC, starting with an overview of the graph HDC setting for node classification. We then introduce the concept of fairness as it applies to GNNs, laying the foundation for understanding its role in our work.

The core idea of HDC-based graph learning is to learn accurate class hypervectors (HVs) based on the data observed during training. These class HVs are then used for classification through predefined metrics, such as computing the similarity between the class HVs and a node embedding. A key advantage of this approach is that it bypasses the need for graph neural network (GNN) matrix operations, simplifying the inference process (e.g., by replacing matrix multiplications with a dot product between HVs). 

\subsection{Notations and Problem Scope} 
\noindent\textbf{Notations.}
Let $\graph=\{\adj,\feat,\sens,\nlabel\}$ denote an attributed graph, where
$\adj\in\mathbb{R}^{N\times N}$ is the adjacency matrix,
$N=|\mathcal{V}|$ is the number of nodes,
$\feat\in{\mathbb{R}^{N\times M}}$ is the node feature matrix,
and $\nlabel\in\{0,\ldots,C-1\}^N$ represents the node labels over $C$ classes.
We use $\sens\in\{0,1\}^{N}$ to denote a binary sensitive attribute
(e.g., gender or race) for each node; extensions to multi-valued attributes
are discussed in the experiments.

In the context of HDC, we define a binary feature vector for each node
$k\in\mathcal{V}$ as
$\vecf_k=[\vf_{k,0}, \vf_{k,1}, \ldots, \vf_{k,M-1}]^\top \in \{0,1\}^M$,
where $M$ is the number of input features.
The binary feature vector $\vecf_k$ is obtained using ID-Level encoding
\citep{imani2017voicehd,gupta2018felix} in this work, but it can also be
generated by other HD encoding schemes, such as non-linear encoding
\citep{imani2020dual}.
HDC-based graph learning aims to obtain a trained HDC model
${\mathcal M}=\{\mathbf{C}_0, \ldots, \mathbf{C}_{C-1}\}$, where
$\mathbf{C}_i \in \mathbb{R}^{1\times D}$ is the class hypervector for class $i$
and $D$ is the hypervector dimensionality.
These class hypervectors aggregate both feature and structural information from
the input graph.
In the following subsection, we detail how feature and structure information are
encoded into hypervectors.

\noindent\textbf{Problem Scope.}
We study group fairness in node classification under the HDC framework.
Given a graph $\graph$, node classification aims to predict the class of
unlabeled nodes based on the graph structure, observed node features, and the
labels of training nodes.
Our goal is to improve the fairness of this classification process by reducing
undesirable statistical dependencies between the model predictions and the
sensitive attribute $\sens$, while preserving the robustness and efficiency
of HDC.
Concretely, we focus on group fairness notions such as demographic parity and
equal opportunity (formalized in \Cref{sec:fair_metric}), and design an HDC
training procedure that mitigates these biases directly at the level of class
hypervectors.
The HDC-based node classification pipeline considered in this work consists of
five main steps:
(1) \emph{Feature Hypervector Encoding},
(2) \emph{Edge Hypervector Encoding},
(3) \emph{Node Hypervector Representation},
(4) \emph{Training of class hypervectors}, and
(5) \emph{Inference}.
In the following sections, we describe each step in detail.

\subsection{Hypervector Representation of Graph HDC}
To obtain the class hypervectors (HVs), we encode the node information from a graph into the high-dimensional (HD) space. Specifically, we map a feature vector, $\vecf$, to a hypervector, $\vecn$, of dimension $\mathbb{R}^{1 \times D}$, where $D$ represents the dimensionality of the HV. To perform this encoding, we define a feature hypervector, $\vecn \in \mathbb{R}^{1 \times D}$, for each node. This is achieved by introducing an orthogonal feature position hypervector, $\vp_i = \{+1, -1\}^D$, for each feature $i$. Here, $D$ is the hypervector's dimension, and for encoding, we utilize a cyclic shift operation, $\rho$, to introduce permutations. 

Graph HDC typically involves three steps: (1) \textit{Feature Hypervector Encoding}, (2) \textit{Edge Hypervector Encoding}, and (3) \textit{Node Hypervector Representation}. It differs from GNN message passing in two key respects: First, graph HDC only requires vector-based operations (e.g., element-wise multiplications), whereas GNNs involve matrix multiplication in forward propagation and backpropagation. Second, graph HDC aims to derive a class hypervector, while GNNs focus on learning network weights. In the following sections, we describe each step of graph HDC in detail.

\subsection{Feature Hypervector Encoding}\label{sec:feat_encode}
We start by randomly generating a base hypervector, $\vp^{\text{base}} \in \{+1, -1\}^D$, and define a matrix of position hypervectors for all features:
\[
\vecp = [\vp_0, \vp_1, \cdots, \vp_{M-1}]^{\top} \in \mathbb{R}^{M \times D},
\]
where each $\vp_i \in \{+1,-1\}^D$ is obtained by applying a cyclic shift $\rho^i$ to $\vp^{\text{base}}$ to encode the position of feature $i$.
For a given node $k$, the binary feature vector is $\vecf_k = [\vf_{k,0}, \vf_{k,1}, \ldots, \vf_{k,M-1}] \in \{0,1\}^M$, where $\vf_{k,i}=1$ indicates the presence of feature $i$ and $\vf_{k,i}=0$ its absence.
The feature hypervector $\vecn_k$ is then computed by accumulating the position hypervectors corresponding to the active features:
\begin{equation}
\begin{split}
\vecn_k
    &= \vecf_k \cdot \vecp^{\top}
     = [\vf_{k,0}, \vf_{k,1}, \cdots , \vf_{k,M-1}]
        \cdot [\vp_0,\vp_1, \cdots ,\vp_{M-1}]^{\top} = \sum_{i=0}^{M-1} \vf_{k,i}\,\vp_i.
\end{split}
\label{eq:define_n}
\end{equation}

\subsection{Edge Hypervector Encoding}\label{sec:edg_encode}
We introduce the concept of an \textit{edge hypervector} for each node, which encodes the node's own hypervector along with the information from its 1-hop and 2-hop neighbors. This encoding is achieved through a hypervector bundling operation. Let $\mathbf{H}^1$ represent the HVs of a node's 1-hop neighbors. To aggregate the neighbor information, we combine the hypervectors of these neighboring nodes.
\begin{equation}
\mathbf{H}^1=\Sigma_{j \in \mathcal{N}_1} \mathbf{N}_j.\label{eq:define_h}
\end{equation}
$\mathcal{N}_1$ denotes the set of indices of a node's 1-hop neighbors. This set can be easily derived from the edge set $E$, which contains the source and destination nodes of each edge.

Similarly, information from 2-hop neighbors can be aggregated to enhance the node representation. Instead of backtracking and searching for 2-hop neighbors, which can be computationally expensive, we take advantage of the previously computed 1-hop neighbor information, $\mathbf{H}^1$, and reuse the 1-hop neighbor set $\mathcal{N}_1$ to construct the 2-hop neighbor representation, $\mathbf{H}^2$. Specifically, for a node, the 2-hop neighbor representation $\mathbf{H}^2$ can be computed as:
\begin{equation}
\mathbf{H}^2=\Sigma_{j \in \mathcal{N}_1} \mathbf{H}_j^1.\label{eq:define_h}
\end{equation}
Here we reuse $\mathcal{N}_1$ from the 1-hop neighbor $\mathrm{HV}\left(\mathbf{H}^1\right)$ computation step.

The encoding and relation embedding stages assign three key properties to each node: $\mathbf{N}$, $\mathbf{H}^1$, and $\mathbf{H}^2$. The previous work RelHD~\citep{kang2022relhd} interprets these as distinct characteristics of a node. 
\subsection{Node Hypervector Representation}
Finally, for each node, we compute the final node hypervector $\mathbf{E}$ as follows:
\begin{align}
\mathbf{E}=\mathbf{N} \odot \phi_0+\mathbf{H}^1 \odot \phi_1+\mathbf{H}^2 \odot \phi_2, \label{eq:define_e}
\end{align}
where $\odot$ indicates the element-wise multiplication (binding operation). Note that $\phi_0, \phi_1$, and $\phi_2$ are bipolar, orthogonal hypervector with $D$ dimensionality, $\{+1,-1\}^D$.

\noindent\textbf{Disucssion.} 
Actually, the entire node representation of the graph HDC is similar to GNNs, but since the calculation is vector multiplication, the graph HDC involves some transformation steps. Specifically, the Feature Hypervector Encoding in~\Cref{sec:feat_encode} is actually the transforming the feature vector to a hypervector, the Edge Hypervector encoding step in~\Cref{sec:edg_encode} corresponds to the neighboring aggregation, and the final node hypervector representation in~\Cref{eq:define_e} corresponds to the message passing. Similar to GNN training, the training of graph HDC algorithms is to use the label information to update the discriminator, which will be used to make predictions. For GNNs, the discriminator is the network weights, and for the graph HDC is the class hypervector.

\subsection{Training}
During the training stage, we construct an HDC model that contains class hypervectors $\mathbf{C}$, where each class is represented by a unique hypervector. To generate the hypervector for the $i$-th class, $\mathbf{C}_i$, we accumulate the edge hypervectors (HVs) of all nodes that belong to class $C_i$. Specifically, the class hypervector $\mathbf{C}_i$ is formed by bundling all the relation HVs of nodes associated with that class.
\begin{align}
\mathbf{C}_i = \sum_{j \in \mathcal{I}_i} \mathbf{E}_j, \ \text{where} \ \mathcal{I}_i=\{\forall j \mid \nlabel_j = i\}, \label{eq:define_c}
\end{align}
where $\mathcal{I}_i$ represents the set of indices of nodes that belong to class $i$. In this case, the number of bits required for class hypervectors (HVs) increases significantly because the number of nodes is much larger than the number of classes. To address the memory overhead and improve inference efficiency, we apply a quantization technique to the HDC model. Specifically, we reduce the model's precision by retaining only the sign bits, a method that has been effectively used in previous HDC-based algorithms~\citep{imani2017voicehd,imani2020dual}.

\subsection{Inference} 
The inference stage is responsible for predicting the label of an unlabeled node by finding the class hypervector (HV) that is most similar to the test data. To do this, we first construct the relation hypervector $\mathbf{E}$ for the test nodes (queries) by reusing the previously computed hypervectors $\mathbf{N}$, $\mathbf{H}^1$, and $\mathbf{H}^2$. For each node, the inference module calculates the similarity between the node’s relation hypervector $\mathbf{E}$ and each class HV. The class with the highest similarity score is then assigned to the node, which can be expressed mathematically as: 
\begin{equation}
\arg \max _i \delta\left(\mathbf{C}_i, \mathbf{E}\right),
\end{equation}
where $\delta$ denotes the similarity function between the class hypervector $\mathbf{C}_i$ and the node’s relation hypervector $\mathbf{E}$.
\begin{equation}
    \hat{\mathbf{Y}} = \underset{i\in\{1,\dots,k\}}{\arg\max} \;\delta\left(\mathrm{Enc}(\mathbf{Y}), \mathbf{C}_i \right),\label{eq:pred_y}
\end{equation}
where $\hat{\mathbf{Y}}$ is the predicted class for $\mathbf{Y}$.

\section{Enabling Fairness in Graph HDC}
In this section, we present our approach {\method} for ensuring fairness in graph HDC. Because the training process of graph HDC differs from that of traditional GNNs, existing debiasing methods cannot be directly applied. To address this, we introduce a bias correction term when updating the class hypervector, thereby removing correlations between the sensitive attribute and the predicted label within the ground-truth label's class hypervector. We begin by defining the fairness metrics and regularizers used in our experiments. We then explain how our framework integrates these components to debias graph HDC.

\subsection{Fairness Concepts and Metrics}\label{sec:fair_metric}
Demographic parity~\citep{dwork2012fairness}, also called statistical parity, requires the prediction $\hat{Y}$ to be independent of the sensitive attribute $S$, i.e., $\hat{Y} \perp S$.
Most of the literature focuses on binary classification and binary attributes, i.e., $Y \in \{0,1\}$ and $S \in \{0,1\}$. In this setting, demographic parity can be written as
\begin{equation}
    P(\hat{Y}=1 \mid S=0) = P(\hat{Y}=1 \mid S=1).
    \label{eq:SP}
\end{equation}
The corresponding demographic parity gap is
\begin{equation}
   \Delta_{DP} = \big|P(\hat{Y}=1\mid S=0)-P(\hat{Y}=1\mid S=1)\big|,
\end{equation}
where a lower $\Delta_{DP}$ indicates a fairer classifier.

Demographic parity naturally extends to multi-class labels and multi-category sensitive attributes by enforcing $\hat{Y} \perp S$.
Following~\citep{locatello2019fairness}, let $Y \in \{Y_1,\dots,Y_c\}$ and $S \in \{S_1,\dots,S_k\}$ denote the class label and the sensitive attribute, where $c$ is the number of classes and $k$ is the number of sensitive groups. We define
\begin{equation}
    \Delta_{DP} = \frac{1}{k} \sum_{i=1}^{k} \max_{j \in \{1,\dots,c\}}
    \big|P(\hat{Y}=Y_j)-P(\hat{Y}=Y_j\mid S=S_i)\big|,
    \label{eq:SP_metric}
\end{equation}
which measures the maximum deviation from independence for each sensitive group, averaged over all groups.
In all our experiments, $\Delta_{DP}$ lies in $[0,1]$ and we report it in percentage points by multiplying the above quantity by $100$.

Equal opportunity~\citep{hardt2016equality} requires that the probability of an instance in the positive class being assigned a positive outcome is equal across sensitive groups, i.e.,
\begin{equation}
     P(\hat{Y}=1\mid Y=1,S=0) = P(\hat{Y}=1\mid Y=1,S=1).
\end{equation}
The corresponding equal opportunity gap is
\begin{equation}
    \Delta_{EO}  =  \big|P(\hat{Y} = 1\mid Y = 1,S = 0)
                     -  P(\hat{Y} = 1\mid Y = 1,S = 1)\big|.
\end{equation}
Similarly, $\Delta_{EO} \in [0,1]$, and we report it in percentage points.

\subsection{Bias Mitigation via Class Hypervector Updates}
Building on the updating strategy used in graph HDC, we design a debiasing mechanism by updating the class hypervectors (HVs).
Specifically, we incorporate the demographic parity gap as a bias correction term $\mathcal{F}$ during the update of the class hypervectors:
\begin{equation}
\mathcal{F} = \alpha \mathcal{B} + \beta,
\label{eq:fair_loss}
\end{equation}
where $\mathcal{F}$ is a scalar fairness regularization term, $\alpha$ and $\beta$ are hyperparameters, and $\mathcal{B}$ is determined by the chosen debiasing method.
In this work, we use a standard gap-based demographic parity regularizer, so $\mathcal{B}$ is computed according to~\Cref{eq:SP_metric}.
In mini-batch training, we estimate $\mathcal{B}$ by averaging the demographic parity gap within each mini-batch using the current predictions.
In practice, we tune $\alpha$ and $\beta$ such that $0 \le \mathcal{F} < 1$ for all mini-batches, so that $\mathcal{F}$ can be interpreted as a shrinkage factor.

The scale of $\mathcal{F}$ is crucial because it directly controls how strongly each mini-batch update is attenuated for debiasing.
An overly large $\mathcal{F}$ can stall learning and harm accuracy, while too small a value yields negligible debiasing.
The hyperparameter $\alpha$ thus plays the role of balancing the influence of $\mathcal{F}$ against the node hypervectors when updating the class hypervectors.
Notably, if $\alpha = 0$ (and $\beta = 0$), there is no debiasing effect and {\method} reduces to the plain graph HDC baseline.

\noindent\textbf{Discussion.}
Unlike traditional GNN training, which optimizes network weights, graph HDC optimizes the class hypervectors.
Consequently, the debiasing step in graph HDC differs from typical GNN-based fairness approaches~\citep{dong2023fairness,chen2024fairness}, where a fairness loss (similar in spirit to $\mathcal{F}$) penalizes the \emph{network weights}.
In graph HDC, this penalty is applied to the \emph{class hypervectors}.
Conceptually, both approaches aim to reduce the statistical dependence between predictions and the sensitive attribute, but our method does so by modulating how strongly each mini-batch can pull the class hypervectors towards potentially biased patterns.
Batches with larger demographic parity gap induce larger $\mathcal{F}$, which shrinks their contribution and prevents the HDC model from overfitting to highly biased training signals.

\medskip
\noindent\textbf{1. The predicted label is correct.}  
When a node $n$ is correctly classified, i.e., $\hat{Y}_n = Y_n$, the standard graph HDC update accumulates the node hypervector $\mathbf{E}_n$ into the class hypervector of the predicted (and ground-truth) label:
\begin{align}
\mathbf{C}_{\hat{Y}_n} \leftarrow \mathbf{C}_{\hat{Y}_n} + \eta\, \mathbf{E}_n,
\end{align}
where $\eta>0$ is the learning rate for class hypervector updates.
We incorporate the bias correction term $\mathcal{F}$ as a \emph{scaling factor} that shrinks the contribution of this mini-batch when the demographic parity gap is large.
The debiased update becomes
\begin{align}
\mathbf{C}_{\hat{Y}_n} \leftarrow \mathbf{C}_{\hat{Y}_n} + \eta\, (1-\mathcal{F})\, \mathbf{E}_n.
\end{align}
Thus, when the current mini-batch is (approximately) fair, $\mathcal{F}\approx 0$ and {\method} behaves like the standard HDC update.
When the mini-batch exhibits a large demographic parity gap, $\mathcal{F}$ grows and $(1-\mathcal{F})$ shrinks, so the class hypervectors move less in response to this biased batch.

\medskip
\noindent\textbf{2. The predicted label is incorrect.}  
When a node is misclassified, i.e., $\hat{Y}_n \neq Y_n$, the graph HDC model updates two class hypervectors: one for the ground-truth label and one for the incorrectly predicted label.
Following the standard HDC learning rule, we add $\mathbf{E}_n$ to the ground-truth class hypervector and subtract $\mathbf{E}_n$ from the predicted class hypervector.
We keep this structure and apply the fairness correction only to the ground-truth class, so that debiasing does not excessively interfere with correcting misclassifications.
Formally,
\begin{align}
\mathbf{C}_{Y_n}      &\leftarrow \mathbf{C}_{Y_n}      + \eta\, (1-\mathcal{F})\, \mathbf{E}_n, \\
\mathbf{C}_{\hat{Y}_n} &\leftarrow \mathbf{C}_{\hat{Y}_n} - \eta\, \mathbf{E}_n.
\end{align}
For the ground-truth label's class hypervector $\mathbf{C}_{Y_n}$, we still \emph{add} the node hypervector $\mathbf{E}_n$ to incorporate this node's information, but we shrink the update by $(1-\mathcal{F})$.
For the wrongly predicted class hypervector $\mathbf{C}_{\hat{Y}_n}$, we \emph{subtract} $\mathbf{E}_n$ without fairness scaling, so that the model can more aggressively correct clearly incorrect prototypes.
This design keeps prediction accuracy as the primary goal when the model is wrong, while still attenuating the influence of highly biased batches on the ground-truth class.

\subsection{Optimization}\label{sec:optimization}
We summarize the entire algorithm in the Appendix.
During training, the feature hypervectors $\vecn$, edge hypervectors $\mathbf{H}^1,\mathbf{H}^2$, and node hypervectors $\mathbf{E}$ are precomputed and remain fixed.
Only the class hypervectors $\mathbf{C}$ are updated according to the rules in the previous subsection, with learning rate $\eta$ and batch-level fairness factor $\mathcal{F}$.

At the beginning of each mini-batch, we use the current class hypervectors $\mathbf{C}$ to obtain predictions $\hat{Y}$ for the nodes in the batch via the similarity-based inference rule in~\Cref{eq:pred_y}.
We set the similarity function $\delta(\cdot,\cdot)$ to cosine similarity between $\ell_2$-normalized hypervectors, which is a standard choice in HDC-based models.
The batch-level demographic parity gap $\mathcal{B}$ is then computed from these predictions and the sensitive attributes using~\Cref{eq:SP_metric}, and the fairness factor $\mathcal{F}=\alpha\mathcal{B}+\beta$ is obtained from~\Cref{eq:fair_loss}.
This scalar $\mathcal{F}$ is shared by all updates within the current mini-batch, producing the fairness-aware scaled updates in the previous subsection.

Unlike traditional GNN training, where optimization is often terminated once the loss
stabilizes below a predefined threshold, our graph HDC training does not rely on a
differentiable loss or an explicit convergence criterion. Instead, following prior HDC
work, we run a small, fixed number of passes over the training data. In all experiments,
we train \emph{both} the vanilla HDC model and {\method} for $20$ epochs, which we found
sufficient for both accuracy and fairness metrics to stabilize, while still using
substantially fewer epochs than the $200$ epochs used for GNN and fairness-aware GNN
baselines. During training, class hypervectors are maintained in full precision, and after
training, we apply sign quantization to $\mathbf{C}$ (keeping only the sign bits) for
memory and inference efficiency, as discussed in~\Cref{eq:define_c}. The full procedure is summarized in~\Cref{alg:training}.

\begin{table}[t]
\centering
\small
\renewcommand{\arraystretch}{1.0}

\begin{tabular}{lcccccc}
\Xhline{1.2pt}
\rowcolor{tablehead!20}
\textbf{Dataset} & \textbf{\#Nodes} & \textbf{\#Edges} & \textbf{\#Feat.} & \textbf{Avg. Degree} & \textbf{Sens.} & \textbf{Label} \\
\hline\hline
\rowcolor{gray!10}{\pokecz} & 67{,}797  & 617{,}958  & 69 & 19.23 & Region      & Working Field \\
{\pokecn} & 66{,}569  & 517{,}047  & 69 & 16.53 & Region      & Working Field \\
\rowcolor{gray!10}{\nba}    & 403       & 10{,}621   & 39 & 53.71 & Nationality & Salary \\
{\german} & 1{,}000   & 21{,}742   & 27 & 44.48 & Gender      & Credit Status \\
\rowcolor{gray!10}{\bail}   & 18{,}876  & 311{,}870  & 18 & 34.04 & Race        & Bail Decision \\
{\credit} & 30{,}000  & 1{,}421{,}858 & 13 & 95.79 & Age       & Future Default \\
\Xhline{1.2pt}
\end{tabular}
\vspace{-0.5em}
\caption{Statistics of commonly used datasets in fair graph learning research (rotated layout).}
\label{tab:dataset-rotated}
\vspace{-1em}
\end{table}

\begin{table*}[t]
\centering
\small
\setlength{\tabcolsep}{2.5pt}
\renewcommand{\arraystretch}{1.05}
\resizebox{\textwidth}{!}{
\begin{tabular}{crcccccccc}
\Xhline{1.2pt}
\rowcolor{tablehead!20}
\multicolumn{1}{c}{\textbf{Dataset}} & \multicolumn{1}{c}{\textbf{Metrics}} &
\textbf{GCN} & \textbf{GraphSAGE} & \textbf{GIN} &
\textbf{FairGNN} & \textbf{NIFTY} & \textbf{EDITS} & \textbf{FairSIN} & {\method} \\
\hline\hline
\rowcolor{gray!10}& ACC($\uparrow$)                        & \underline{\ms{69.33}{0.72}}  & {\bf \ms{70.49}{0.86}}       & \ms{68.54}{0.91}            & \ms{64.55}{0.68}            & \ms{63.67}{0.91}   & OOM             & \ms{67.44}{0.73}             & \ms{67.45}{0.42}   \\
& F1($\uparrow$)                         & \ms{67.65}{0.94}              & {\bf \ms{69.52}{0.99}}       &\underline{\ms{67.97}{0.93}} & \ms{65.04}{0.74}            & \ms{64.42}{0.82}   & OOM             & \ms{65.35}{0.86}             & \ms{65.57}{0.51}   \\
\rowcolor{gray!10}& $\Delta_{\mathit{DP}}$ ($\downarrow$)  & \ms{5.07}{2.27}               & \ms{4.67}{2.93}              & \ms{4.19}{2.25}             & \ms{3.39}{1.93}             & \ms{4.94}{2.66}    & OOM             &\underline{\ms{2.67}{2.28}}   & {\bf \ms{2.85}{2.24}}    \\ 
\multirow{-4}{*}{\pokecz}
& $\Delta_{\mathit{EO}}$ ($\downarrow$)  & \ms{5.44}{1.79}               & \ms{3.37}{1.76}              & \ms{4.06}{1.83}             & \ms{2.52}{1.03}             & \ms{3.53}{1.88}    & OOM             &\underline{\ms{3.08}{1.27}}   & {\bf \ms{3.92}{1.35}}    \\ \hline
\rowcolor{gray!10}& ACC($\uparrow$)                        & {\bf \ms{70.76}{1.34}}        & \ms{69.23}{1.69}             &\underline{\ms{69.73}{2.25}} & \ms{64.64}{2.02}            & \ms{62.36}{2.55}   & OOM             & \ms{66.21}{2.20}             & \ms{66.25}{2.66}   \\
& F1($\uparrow$)                         & \ms{66.53}{0.95}              &\underline{\ms{68.66}{0.49}}  & {\bf \ms{68.86}{2.04}}      & \ms{62.31}{1.54}            & \ms{60.76}{1.17}   & OOM             & \ms{64.36}{1.37}             & \ms{63.38}{1.89}   \\
\rowcolor{gray!10}& $\Delta_{\mathit{DP}}$ ($\downarrow$)  & \ms{8.44}{2.49}               & \ms{6.87}{2.10}              & \ms{5.42}{5.87}             & \ms{6.77}{1.73}             & \ms{5.62}{1.73}    & OOM             &\underline{\ms{2.68}{1.79}}   & {\bf \ms{2.36}{2.77}}    \\ 
\multirow{-4}{*}{\pokecn}
& $\Delta_{\mathit{EO}}$ ($\downarrow$)  & \ms{12.23}{2.80}              & \ms{10.39}{1.94}             & \ms{9.57}{4.65}             & \ms{8.98}{2.82}             & \ms{7.90}{2.06}    & OOM             &\underline{\ms{3.92}{1.42}}   & {\bf \ms{3.42}{1.43}}    \\  \hline
\rowcolor{gray!10}& ACC($\uparrow$)                        &\underline{\ms{72.78}{1.13}}  & {\bf \ms{73.04}{1.45}}        & \ms{70.89}{1,76}            & \ms{70.45}{1.65}            & \ms{62.37}{1.63}   & \ms{63.54}{2.33}            & \ms{68.14}{2.11}             & \ms{70.87}{1.81}   \\
& F1($\uparrow$)                         &\underline{\ms{74.01}{1.24}}  & {\bf \ms{75.22}{0.69}}        & \ms{73.03}{1.34}            & \ms{73.54}{0.95}            & \ms{66.44}{1.86}   & \ms{67.31}{1.56}            & \ms{69.35}{1.15}             & \ms{73.93}{1.20}   \\
\rowcolor{gray!10}& $\Delta_{\mathit{DP}}$ ($\downarrow$)  & \ms{2.99}{2.17}              & \ms{5.33}{2.91}               & \ms{3.41}{2.44}             & \ms{2.78}{1.23}             & \ms{6.67}{3.17}    & \ms{2.82}{2.88}             &\underline{\ms{2.67}{1.26}}   & {\bf  \ms{2.80}{1.64}}    \\ 
\multirow{-4}{*}{\nba}
& $\Delta_{\mathit{EO}}$ ($\downarrow$)  & \ms{4.06}{2.93}              & \ms{5.40}{3.22}               & \ms{4.67}{1.76}             & \ms{2.23}{1.56}             & \ms{4.23}{2.22}    &\underline{\ms{1.60}{1.51}}  & \ms{3.25}{1.45}              & {\bf \ms{3.92}{1.34}}    \\ \hline
\rowcolor{gray!10}& ACC($\uparrow$)                        & {\bf \ms{71.35}{1.22}}       & \ms{69.15}{1.55}              & \ms{69.22}{1.64}            & \ms{69.23}{1.63}            & \ms{68.93}{1.68}   & \ms{70.17}{2.84}            & \ms{69.52}{1.87}             &\underline{\ms{70.30}{1.73}}    \\
& F1($\uparrow$)                         & \ms{79.47}{2.10}             & {\bf  \ms{79.80}{0.89}}       & \ms{77.38}{1.23}            & \ms{78.57}{1.46}            & \ms{79.85}{1.83}   & \ms{80.54}{1.53}            &\underline{\ms{79.64}{1.54}}  & \ms{79.25}{1.29}    \\
\rowcolor{gray!10}& $\Delta_{\mathit{DP}}$ ($\downarrow$)  & \ms{37.85}{16.02}            & \ms{30.37}{11.64}             & \ms{36.66}{14.67}           & \ms{3.78}{1.96}             & \ms{4.62}{2.14}    &\underline{\ms{2.86}{2.87}}  & \ms{2.76}{1.75}              & {\bf \ms{2.57}{1.75}}     \\ 
\multirow{-4}{*}{\german}
& $\Delta_{\mathit{EO}}$ ($\downarrow$)  & \ms{29.24}{15.16}            & \ms{19.53}{10.88}             & \ms{27.85}{19.54}           & \ms{4.04}{2.41}             & \ms{3.74}{1.35}    & \ms{4.98}{1.49}             &\underline{\ms{3.78}{1.59}}   & {\bf \ms{3.07}{1.56}}     \\ \hline
\rowcolor{gray!10}& ACC($\uparrow$)                        & \ms{84.27}{1.57}             & {\bf \ms{85.71}{1.62}}        & \ms{83.72}{1.07}            &\underline{\ms{84.65}{2.52}} & \ms{77.02}{1.66}   & \ms{81.87}{2.36}            & \ms{84.16}{0.78}             & \ms{83.42}{1.77}     \\
& F1($\uparrow$)                         & \ms{79.05}{1.73}             & {\bf \ms{80.67}{0.97}}        & \ms{78.15}{1.88}            & \ms{78.32}{1.37}            & \ms{76.71}{1.73}   & \ms{75.42}{1.69}            &\underline{\ms{79.46}{1.79}}  & \ms{80.81}{1.32}     \\
\rowcolor{gray!10}& $\Delta_{\mathit{DP}}$ ($\downarrow$)  & \ms{8.96}{5.54}              & \ms{7.23}{4.61}               & \ms{6.48}{4.43}             & \ms{7.08}{4.21}             & \ms{5.99}{2.15}    & \ms{6.65}{2.83}             &\underline{\ms{2.73}{0.73}}   & {\bf \ms{2.37}{2.64}}      \\ 
\multirow{-4}{*}{\bail}
& $\Delta_{\mathit{EO}}$ ($\downarrow$)  & \ms{5.66}{2.16}              & \ms{6.89}{3.72}               & \ms{4.65}{1.86}             & \ms{5.13}{2.16}             & \ms{5.80}{2.45}    & \ms{9.49}{1.21}             &\underline{\ms{4.25}{1.35}}   & {\bf \ms{3.55}{1.36}}      \\\hline
\rowcolor{gray!10}& ACC($\uparrow$)                        &\underline{\ms{73.46}{1.02}}  & \ms{69.24}{1.66}              & {\bf \ms{73.79}{2.05}}      & \ms{70.53}{2.63}            & \ms{71.65}{1.76}   & \ms{71.53}{2.28}            & \ms{69.87}{1.96}             & \ms{70.84}{2.55}    \\
& F1($\uparrow$)                         & {\bf \ms{82.88}{1.64}}       & \ms{67.66}{0.89}              & \ms{81.53}{1.72}            & \ms{80.26}{1.63}            & \ms{80.73}{1.33}   &\underline{\ms{81.56}{1.17}} & \ms{82.78}{1.25}             & \ms{81.22}{1.83}    \\
\rowcolor{gray!10}& $\Delta_{\mathit{DP}}$ ($\downarrow$)  & \ms{12.30}{6.51}             & \ms{14.58}{7.97}              & \ms{11.85}{5.37}            & \ms{6.86}{1.26}             & \ms{8.80}{2.08}    & \ms{9.14}{2.25}             &{\bf \ms{2.19}{0.76}}         & \underline{\ms{2.97}{2.08}}     \\ 
\multirow{-4}{*}{\credit}
& $\Delta_{\mathit{EO}}$ ($\downarrow$)  & \ms{9.93}{7.84}              & \ms{10.79}{5.49}              & \ms{8.76}{4.69}             & \ms{7.45}{3.83}             & \ms{6.98}{3.76}    & \ms{7.35}{1.45}             &\underline{\ms{3.65}{1.15}}   & {\bf \ms{3.04}{1.95}}     \\ \Xhline{1.2pt}
\end{tabular}}
\vspace{-0.5em}
\caption{Fairness Evaluation of Existing Models on Graph Datasets Using Race and Gender as Sensitive Attributes. An upward arrow ($\uparrow$) indicates that higher values are better, while a downward arrow ($\downarrow$) indicates the opposite. Results are averaged over $10$ runs per method, with the best results in \textbf{bold} and the runner-up results \underline{underlined}. OOM means out of memory.}
\label{tab:acc_fair}
\vspace{-1em}
\end{table*}

\section{Experiments}
\noindent\textbf{Datasets, Baselines, and Experimental Setups.} We conduct experiments using six fairness graph datasets from various domains, each with different sensitive attributes: {\german}, {\bail}, {\credit}, {\pokecn}, {\pokecz}, and {\nba}. More details about datasets can be found in~\Cref{app:dataset}. To evaluate the performance of {\method}, we compare it against three standard
GNN models: GCN~\citep{gcn}, GraphSAGE~\citep{sage}, and GIN~\citep{gin}---and
four state-of-the-art fairness-aware GNN methods: FairGNN~\citep{fairgnn},
NIFTY~\citep{nifty}, EDITS~\citep{edits}, and FairSIN~\citep{yang2024fairsin}.
A detailed introduction to these baseline models is provided in
\Cref{app:baseline}.
For a fair and controlled comparison, we use two-layer GCN/GraphSAGE/GIN
architectures with 128 hidden units and a learning rate of $10^{-2}$ for all
GNN-based methods.
For fairness-aware GNNs that wrap a GNN encoder (FairGNN, EDITS, and FairSIN), we follow the common practice in their original works and adopt a two-layer GCN
with 128 hidden units as the backbone encoder, while NIFTY uses a one-layer GCN
as in the original paper.
Our HDC-based method {\method} does not rely on a GNN backbone; instead, it
operates in a hyperdimensional space with hypervector dimensionality
$D = 4{,}096$.
Training schedules follow the optimization setup described in
\Cref{sec:optimization} ($200$ epochs with early stopping for GNN/fairness-aware
GNN baselines, and $20$ epochs for HDC and {\method}).

\subsection{How Does {\method} Perform on Fairness and Node Classification Accuracy?}
\begin{figure}[t]
\centering
\includegraphics[width=1.0\linewidth]{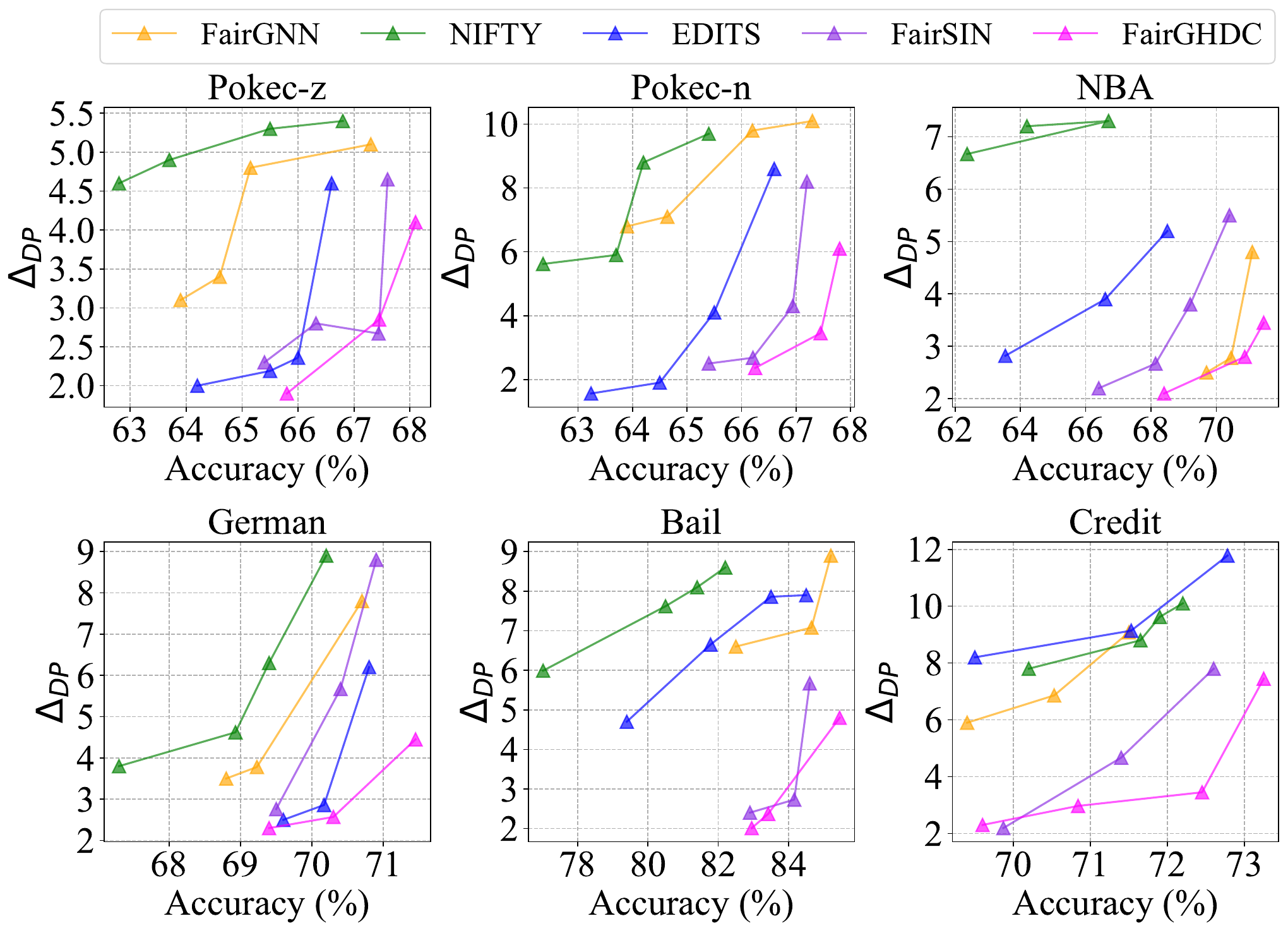}
\vspace{-2em}
\caption{Trade-off between fairness and node classification accuracy across six datasets. Results in the upper left corner, which exhibit lower bias and higher accuracy, represent the ideal balance.} \label{fig:tradeoff}
\vspace{-1em}
\end{figure}
We compare the proposed {\method} with standard and fairness-aware GNNs across six datasets. For each dataset, we performed $10$ different splits to ensure robustness, calculating the mean and standard deviation for each metric across these splits. The accuracy and fairness performance of node classification is presented in \Cref{tab:acc_fair}. The results show that:
\begin{fullitemize}
\item {\method} typically achieves the lowest $\triangle_{DP}$ and $\Delta_{\mathit{EO}}$, except on {\credit}, where FairSIN achieves the lowest $\Delta_{\mathit{DP}}$. Notably, due to the large scale of the Pokec datasets~\Cref{tab:dataset-rotated}, EDITS ran out of memory. 
\item {\method} demonstrates strong fairness performance with minimal utility loss, ranking first or second for ACC and F1 scores among fairness-aware GNNs. On {\german}, {\method} even outperforms GraphSAGE and GIN in ACC while maintaining the lowest $\Delta_{\mathit{DP}}$ and $\Delta_{\mathit{EO}}$. 
\item On {\pokecn}, where all fairness-aware GNNs show a larger ACC gap compared to standard GNNs, our method still achieves the highest ACC among fairness-aware GNNs.
\end{fullitemize}

\subsection{What Is the Fairness–Utility Trade-Off for {\method} and the Baselines?}
We evaluate the trade-off between accuracy and $\Delta_{DP}$ for the baseline methods by varying their fairness hyperparameters~\citep{yao2023stochastic,fairdrop}. In~\Cref{fig:tradeoff}, different colors are used to distinguish FairGNN, NIFTY, EDITS, FairSIN, and {\method}. Ideally, a debiasing method should be located in the upper-left corner of the plot, representing an optimal balance between utility and fairness. As shown in the figure:
\begin{fullitemize}
\item {\method} generally offers the most favorable trade-offs between accuracy and fairness ($\Delta_{DP}$). In contrast, EDITS tends to achieve stronger debiasing but at the cost of greater utility loss, while FairSIN is better at maintaining high utility but is less effective at reducing bias. 
\item When comparing to the results in~\Cref{tab:acc_fair}, we observe that sometimes the $\Delta_{\mathit{EO}}$ values are higher, even compared to standard GNNs. This occurs because the emphasis on fairness in these methods often leads to a less optimized utility outcome.
\end{fullitemize}

\subsection{How Does {\method}'s Efficiency Compare to Baseline Methods?}
\begin{figure}[t]
\centering
\includegraphics[width=1.0\linewidth]{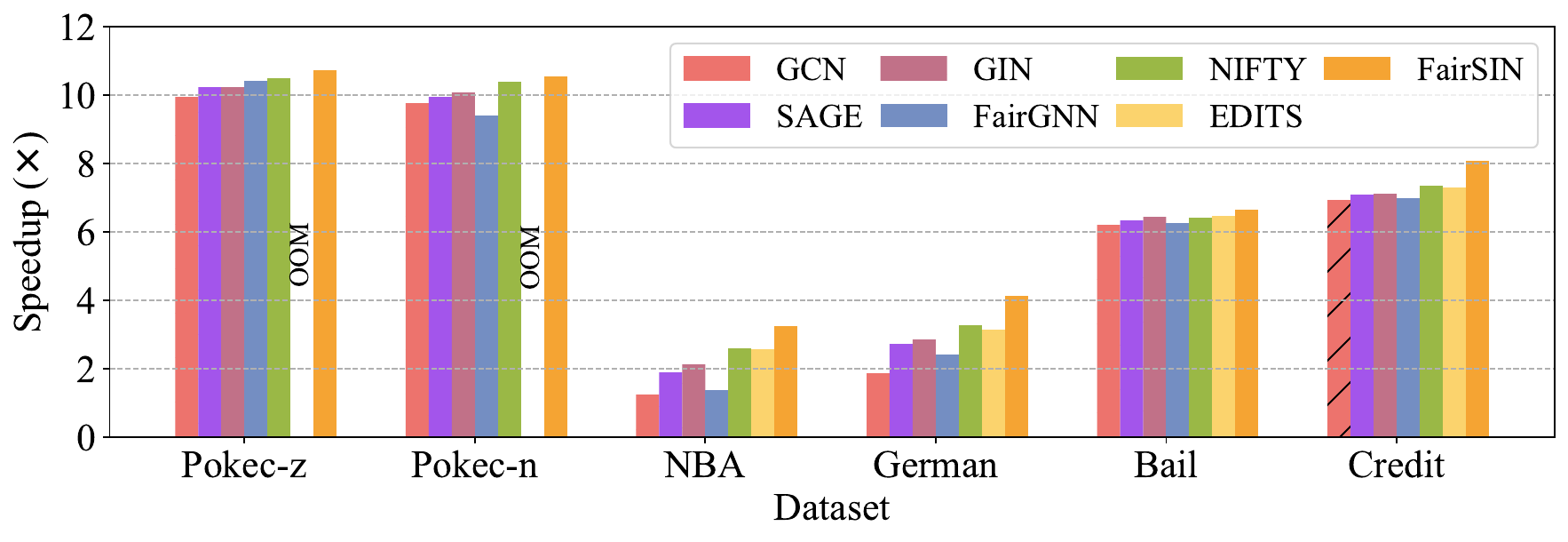}
\vspace{-2em}
\caption{Runtime performance improvement of {\method} over baseline methods on GPU.
Bars show the speedup (runtime(baseline)/runtime({\method})) on each dataset.
For the {\pokecn} and {\pokecz} datasets, EDITS encountered an out-of-memory (OOM)
issue. The corresponding wall-clock training times (in seconds) are reported in
\Cref{tab:hdc_runtime_seconds}. Slashed bars $\slashbox$ represent general GNN methods,
while blank bars $\blankbox$ represent fairness-aware GNN methods.}~\label{fig:runtime}
\vspace{-2em}
\end{figure}
We implement {\method} and baseline methods on an NVIDIA GeForce RTX 4090 with CUDA v12.4. GPU power consumption was measured using `nvidia-smi`. The statistics of the graph datasets used in our experiments are provided in~\Cref{tab:dataset-rotated}. {\pokecn}, {\pokecz}, and {\credit} are relatively large graphs, while {\nba}, {\german}, and {\bail} are smaller datasets. To ensure a fair comparison of training times, we included the execution time for encoding, relation embedding, training, and retraining phases in {\method}, as baseline algorithms also include feature extraction during their training steps. 

From the results in~\Cref{fig:runtime}, we have the following observations:
\begin{fullitemize}
    \item On the large {\pokecz} dataset, {\method} outperforms \textit{general GNN methods} (marked by $\slashbox$) with speedups of about $10\times$ over GCN, GraphSAGE, and GIN.
    \item On the similarly large {\pokecn} dataset, {\method} surpasses \textit{fairness-aware GNNs} (marked by $\blankbox$) by roughly $9\text{--}11\times$ for FairGNN, NIFTY, and FairSIN. These large graphs are challenging to train efficiently, and EDITS encounters an out-of-memory (OOM) error on both {\pokecz} and {\pokecn}.
    \item On smaller datasets, the gains are more moderate but still non-trivial. On {\nba}, {\method} is about $1.2\times$ faster than GCN and $3.3\times$ faster than FairSIN; on {\german}, {\bail}, and {\credit}, {\method} achieves between roughly $2\times$ and $8\times$ speedup over GNN and fairness-aware GNN baselines.
\end{fullitemize}

\subsection{How Do Hyperparameters Affect {\method}’s Performance?}
\begin{figure}[t]
\centering
\hspace{-1em}
\includegraphics[width=0.51\linewidth]{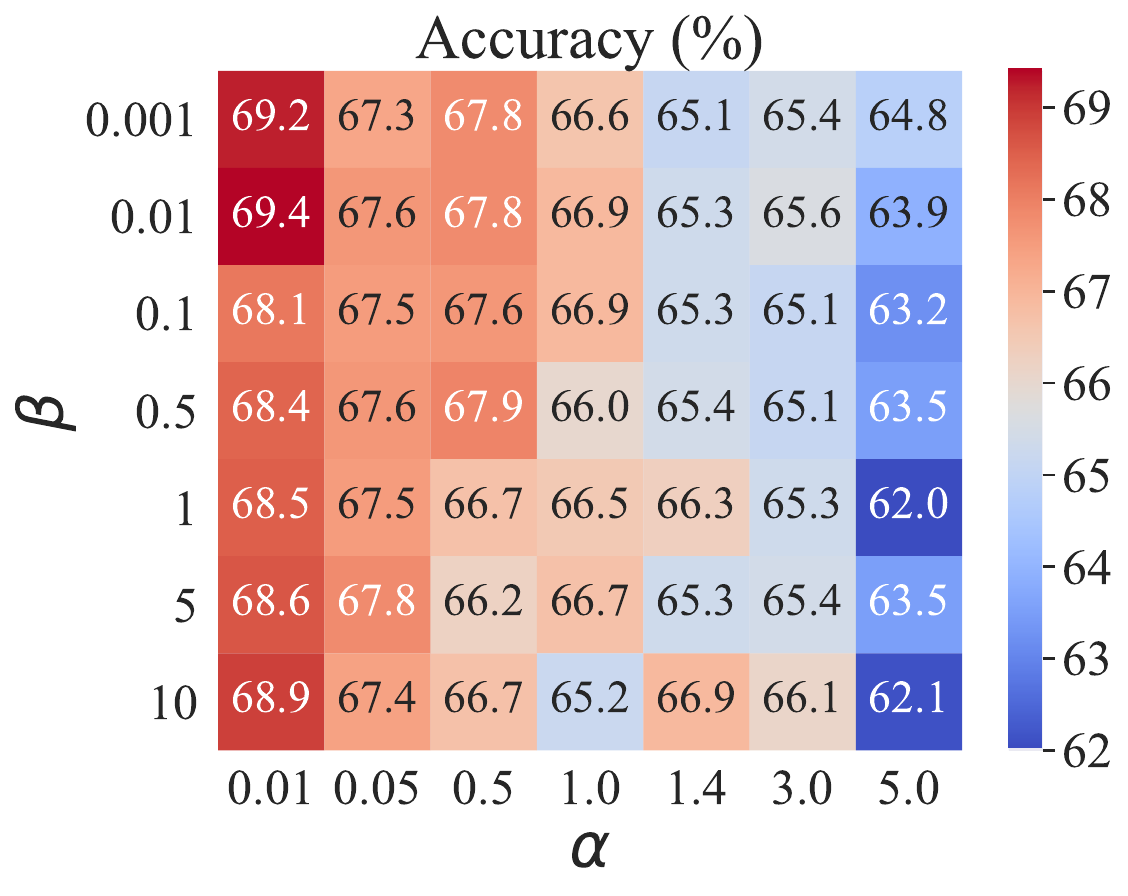}\hspace{-1em}
\includegraphics[width=0.47\linewidth]{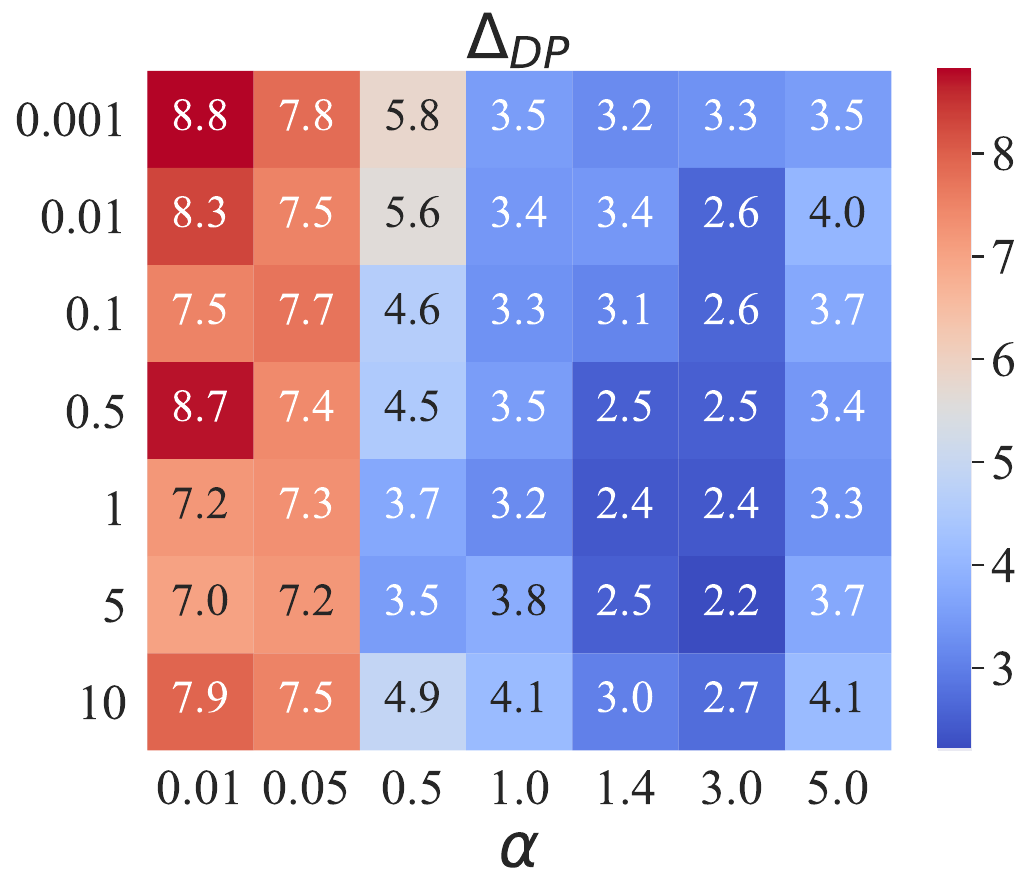}\\
\hspace{-1em}
\includegraphics[width=0.51\linewidth]{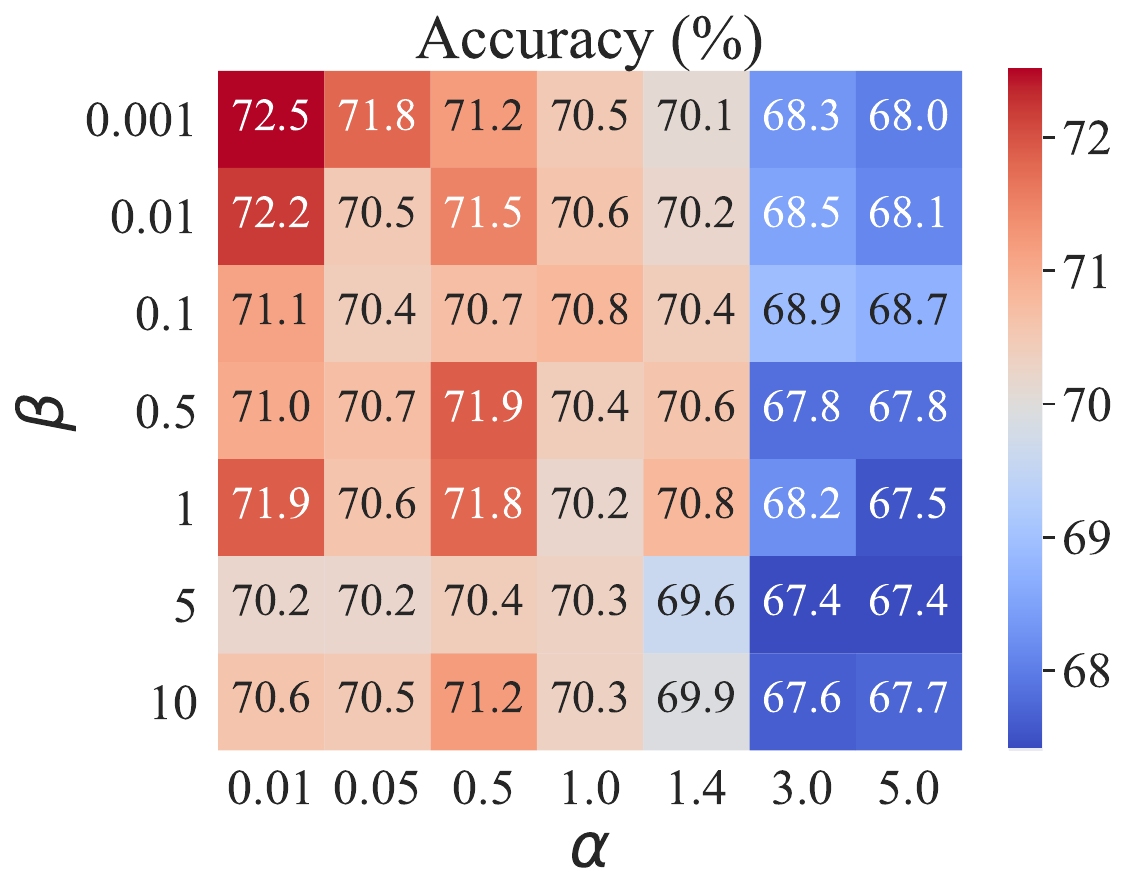}\hspace{-1em}
\includegraphics[width=0.47\linewidth]{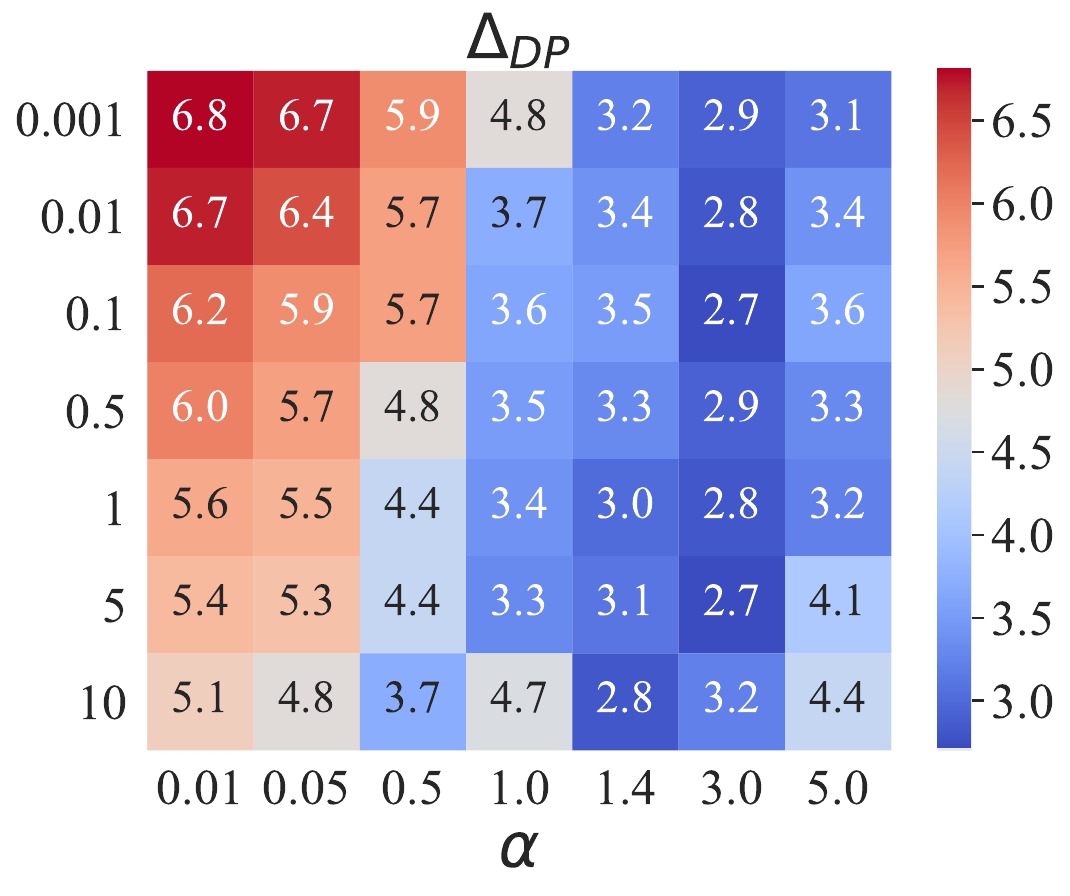}
\vspace{-1em}
\caption{Parameter sensitivity analysis on {\pokecn} (first row) and {\credit} (second row).} \label{fig:cs_hyper}
\vspace{-1em}
\end{figure}
For all models, we tune hyperparameters on the training set using cross-validation and select the optimal configuration via grid search. In particular, for {\method} we vary the parameters $\alpha$ and $\beta$ over the ranges $[10^{-2}, 1.0]$ and $[10^{-3}, 10]$, respectively. The heatmaps in \Cref{fig:cs_hyper} visualize the accuracy and $\Delta_{DP}$ achieved by different $(\alpha,\beta)$ combinations on the {\pokecn} and {\credit} datasets. In the accuracy plots, warmer colors indicate higher values (better utility), whereas in the $\Delta_{DP}$ plots, cooler colors correspond to lower disparity (better fairness). The results show that:
\begin{fullitemize}
\item The highest accuracy for both {\pokecn} and {\credit} is achieved when $\alpha$ is set to the smallest value, $1e-3$, while the best fairness performance occurs when $\alpha = 1.4$. 
\item The fairness performance is more sensitive to variations in $\alpha$ than in $\beta$. For instance, increasing $\beta$ from $1e-3$ to $10$ (a $10,000\times$ increase) results in only a slight improvement in $\triangle_{DP}$, from $8.8$ to $7.9$ for {\pokecn} and from $6.8$ to $5.1$ for {\credit}. 
\item Increasing $\alpha$ from $0.05$ to $0.5$ (a $10\times$ increase) leads to a significant drop in $\triangle_{DP}$ from $5.8$ to $3.5$ for {\pokecn}, and from $5.9$ to $4.8$ for {\credit}, while keeping $\beta$ fixed at $1e-3$.
\end{fullitemize}

\vspace{-1em}
\section{Conclusion}
\vspace{-0.7em}
In this paper, we take a first step toward improving fairness in graph 
hyperdimensional computing (HDC) by introducing a simple 
fairness-aware training mechanism that reduces bias while 
preserving the robustness and efficiency of graph HDC. Our approach incorporates a bias correction term, derived from a demographic-parity regularizer, into the update of class hypervectors: for each sample, we use a scalar fairness factor to scale the contribution of its node hypervector to the class hypervector of the ground-truth label, while keeping the update rule for the predicted label focused on accuracy. This allows {\method} to promote fair decision-making directly in the hypervector space without modifying the encoder or relying on backpropagation. Empirically, {\method} achieves substantially lower demographic-parity and equal-opportunity gaps than standard HDC and several GNN-based baselines, while maintaining accuracy comparable to both standard and fairness-aware GNNs. At the same time, {\method} retains the computational advantages of HDC, yielding between roughly $1.1\times$ and $10.7\times$ speedup in training time on GPU compared to GNN and fairness-aware GNN methods across six benchmark datasets. These results highlight the scalability and fairness benefits of our approach for large-scale graph-based tasks.

\section*{Acknowledgements}
This work was supported in part by the DARPA Young Faculty Award, the National Science Foundation (NSF) under Grants \#2127780, \#2319198, \#2321840, \#2312517, and \#2235472, \#2431561, the Semiconductor Research Corporation (SRC), the Office of Naval Research through the Young Investigator Program Award, and Grants \#N00014-21-1-2225 and \#N00014-22-1-2067, Army Research Office Grant \#W911NF2410360. Additionally, support was provided by the Air Force Office of Scientific Research under Award \#FA9550-22-1-0253, along with generous gifts from Xilinx and Cisco.

\bibliographystyle{plain}
\bibliography{ref}

\clearpage
\appendix
\section{Related Work}\label{sec:related_work}
We begin by reviewing relevant studies, focusing on recent developments in Graph Neural Networks and Graph Hyperdimensional Computing. We then examine the existing algorithms for fairness-aware GNNs, including the baselines used in our experiments.
\subsection{Graph Neural Networks} 
Graph-structured data is prevalent in real-world applications like social networks and biological systems. Numerous graph neural networks (GNNs) have been proposed to effectively model this non-Euclidean data~\citep{wu2020comprehensive,zhou2020graph,yuan2020explainability,liu2025beyond}. GNNs learn meaningful representations of nodes, edges, and entire graphs for various tasks. For example, the Graph Convolutional Network (GCN)~\citep{gcn} aggregates and refines node features through convolutional operations. The Graph Isomorphism Network (GIN)~\citep{gin} increases GNN expressiveness, matching the power of the Weisfeiler-Lehman (WL) test~\citep{shervashidze2011weisfeiler} for graph isomorphism. GraphSAGE~\citep{sage} introduces inductive learning for scalability on large graphs, while the Graph Attention Network (GAT)~\citep{gat} uses attention mechanisms to weigh neighbors during message passing. These GNN variants have generally shown strong performance in many graph-based tasks.

\subsection{Graph Hyperdimensional Computing}
Several recent studies have explored representing graph data in hyperdimensional (HD) space for graph-based machine learning tasks. GrapHD~\citep{poduval2022graphd} encodes node and neighbor information into a single hypervector, which can be used to reconstruct the original graph and aid tasks like object detection when combined with CNN algorithms. Additionally, they proposed a PIM-based hardware accelerator to speed up processing. However, GrapHD lacks support for on-device training and inference, and it only operates on binary HVs, which can lead to data loss. In~\citep{nunes2022graphhd}, graphs are encoded by bundling edge hypervectors, derived from binding node HVs. While this approach supports graph classification using HD computing principles, it overlooks the structural role of nodes and is limited to GPU implementation, despite HD computing's advantages in speed and energy efficiency on processing in-memory-based hardware~\citep{wu2018brain,li2016hyperdimensional,gupta2018felix}. RelHD~\citep{kang2022relhd} captures node relationships and integrates them into an HD-based training algorithm, considering node structure and enabling on-device graph-based machine learning.

\subsection{Algorithmic Fairness in GNNs} \label{sec:rel_fairgnn}
Fairness in machine learning has been widely studied and has attracted broad attention due to its societal importance, spanning applications from vision and graphs to large language models~\citep{wan2023processing,liu2025white,le2022survey}.
In Graph Neural Networks, various fairness definitions have been proposed~\citep{dp,eo,individual,cf,degree,wang2022uncovering}, with group fairness being one of the most prominent~\citep{dp,eo}. Group fairness ensures that predictions are unbiased across demographic groups~\citep{liu2025enablingcikm,liucauchy}. Recent state-of-the-art methods~\citep{liu2024promoting,liu2025enablingfgu,liu2023fairgraph} for improving group fairness in GNNs focus on either feature masking or topology modification during message passing. FairGNN~\citep{fairgnn} uses a GCN-based estimator to predict sensitive attributes for nodes with missing data and incorporates adversarial loss based on both estimated and adversary-predicted sensitive attributes. NIFTY~\citep{nifty} enforces counterfactual fairness through feature perturbation and edge dropping, while EDITS~\citep{edits} reduces discrimination by pruning graph topology and node features. FairSIN~\citep{yang2024fairsin} employs a \textit{neutralization-based} strategy by introducing \textit{Fairness-facilitating Features} to node representations before message passing. FairVGNN~\citep{fairv} applies a mask generator to remove channels correlated with sensitive attributes. FairDrop~\citep{fairdrop} uses edge masking to counteract homophily. Some approaches~\citep{edits,ling2023learning} not only drop biased edges but also add new edges to improve fairness.

\section{Introduction to Datasets and Baselines}
\subsection{Datasets}\label{app:dataset}
\begin{fullitemize}
\item {\german}: This dataset models credit clients as nodes, where each node is associated with demographic and financial attributes as well as a binary credit-risk label. 
Edges are constructed between clients with similar profiles (e.g., age, income, and account status), following prior work on graph-based fairness.

\item {\bail}: Nodes represent defendants released on bail between $1990$ and $2009$, with node features capturing criminal history, demographics, and case-related variables. 
Edges connect defendants with similar criminal records and demographic attributes, yielding a graph where the node label indicates whether the defendant re-offended while on release.

\item {\credit}: In this dataset, nodes correspond to credit card users and are annotated with features describing their financial status (e.g., credit limit, repayment history) and a binary default label. 
We build edges between users whose purchasing and payment behaviors are similar, creating a credit risk prediction graph used in prior fairness studies.

\item Pokec: This Slovak online social network, anonymized in $2012$, contains user profiles with rich demographic and interest attributes. 
Following standard practice, we construct two subgraphs, {\pokecn} and {\pokecz}, corresponding to users from different Slovak regions; nodes are users, edges are friendships, and labels capture user engagement outcomes, while sensitive attributes come from profile fields such as gender.

\item {\nba}: This dataset contains statistics for approximately $400$ professional {\nba} players, including performance, salary, and biographical features. 
We connect players with similar on-court statistics to form the graph, and use a binary label derived from career outcomes; the sensitive attribute is the player’s nationality (U.S. vs. non-U.S.), as in prior fairness work.
\end{fullitemize}

\subsection{Baseline Methods}\label{app:baseline}
We compare {\method} with both standard GNN architectures and state-of-the-art
fairness-aware GNN methods:

\begin{fullitemize}
\item \textbf{GCN}~\citep{kipf2017semi}.
Graph Convolutional Networks (GCN) are a canonical message-passing GNN
architecture that aggregates and linearly transforms the features of
neighboring nodes, followed by non-linear activations.
GCN serves as a strong accuracy-oriented baseline without any explicit
fairness regularization.

\item \textbf{GraphSAGE}~\citep{hamilton2017inductive}.
GraphSAGE is an inductive GNN that learns aggregation functions (e.g.,
mean, max-pooling) to combine information from sampled neighbors.
It is widely used in practice for large graphs and, like GCN, optimizes
only for predictive performance.

\item \textbf{GIN}~\citep{gin}.
The Graph Isomorphism Network (GIN) is designed to match the expressive
power of the Weisfeiler–Lehman test by using sum aggregation and
multi-layer perceptrons.
GIN often achieves strong accuracy on node classification tasks and is
used here as another accuracy-focused backbone.

\item \textbf{FairGNN}~\citep{fairgnn}.
FairGNN introduces an adversarial debiasing framework on graphs:
a primary GNN predicts labels, while an auxiliary adversary tries to
predict the sensitive attribute from node representations.
By jointly training the predictor and adversary, FairGNN aims to
reduce the dependence between predictions and sensitive attributes,
even when sensitive labels are partially observed.

\item \textbf{NIFTY}~\citep{nifty}.
NIFTY (uNIfying Fairness and stabiliTY) is a framework for learning
node representations that are both fair and stable.
It leverages augmented graphs and a novel objective that promotes
similarity between representations of the original and the counterfactual
nodes, together with layer-wise weight normalization, to encourage
counterfactual fairness and robustness.

\item \textbf{EDITS}~\citep{edits}.
EDITS is a model-agnostic, data-centric debiasing framework that
explicitly models bias in attributed networks and edits the graph
(topology and/or node attributes) before GNN training.
By optimizing bias-aware objectives and producing a less biased input
graph, EDITS can improve fairness for a variety of downstream GNN
models.

\item \textbf{FairSIN}~\citep{yang2024fairsin}.
FairSIN (Sensitive Information Neutralization) adopts a neutralization-based strategy: it introduces additional fairness-facilitating features (F3)
derived from heterogeneous neighbors to statistically neutralize
sensitive bias in node representations while preserving useful
non-sensitive information.
It provides both data-centric and model-centric variants for improving
group fairness in GNNs.
\end{fullitemize}

\section{Time Complexity Analysis}
We briefly compare the time complexity of {\method} with that of a
representative fairness-aware GNN, FairGNN.

\subsection{{\method}}
Let $N$ be the number of nodes, $E$ the number of edges, $C$ the number of
classes, $D$ the hypervector dimension, and $T$ the number of training epochs.
The HDC pipeline has two main parts.

\emph{Encoding.}
Feature, edge, and node hypervector construction involve only vector-wise
operations (e.g., additions and element-wise multiplications).
Given $D$-dimensional hypervectors, these steps cost
$\mathcal{O}((N+E)\cdot D)$ overall and are performed once before training.

\emph{Training and inference.}
During training, for each node, {\method} performs:
(i) a relation-based hypervector lookup,
(ii) similarity computation with each of the $C$ class hypervectors
(e.g., cosine similarity), and
(iii) an update of the corresponding class hypervector(s).
Each of these operations is linear in $D$, so the per-node cost is
$\mathcal{O}(C \cdot D)$ and one training epoch costs
$\mathcal{O}(N \cdot C \cdot D)$.
Over $T$ epochs, the training complexity is therefore
$\mathcal{O}(T \cdot N \cdot C \cdot D)$.
Inference reuses the precomputed node hypervectors and only evaluates the
similarity to the $C$ class hypervectors, which costs
$\mathcal{O}(N \cdot C \cdot D)$.

The fairness regularizer in {\method} is based on a gap-style demographic-parity
measure computed over mini-batches.
Its cost is $\mathcal{O}(N)$ per epoch (linear in the number of examples and
constant in $D$ and $C$), which is asymptotically dominated by the
$\mathcal{O}(N \cdot C \cdot D)$ similarity and update operations.
In our experiments $C$ and $D$ are fixed and small (e.g., $C \le 2$,
$D = 4{,}096$), so the overall complexity of {\method} scales essentially
linearly with the number of nodes.

\subsection{FairGNN}
Let $F$ be the number of input features, $H$ the hidden dimension size, $L$
the number of GNN layers, and $T$ the number of training epochs.
FairGNN consists of a GNN-based graph encoder, a label classifier, and a
sensitive-attribute predictor trained in an adversarial manner.
For a standard message-passing GNN layer, the forward (and backward) pass has
complexity $\mathcal{O}((E+N)\cdot H)$; stacking $L$ layers yields
$\mathcal{O}(L\cdot(E+N)\cdot H)$ per epoch for the encoder.
The label classifier adds $\mathcal{O}(N \cdot H \cdot C)$ per epoch.
Because FairGNN uses an adversarial debiasing scheme, the encoder and classifier
parameters are updated both for the prediction loss and for the adversarial loss,
effectively doubling the backpropagation cost.
Ignoring constant factors, the total time complexity of FairGNN over $T$ epochs
is $\mathcal{O}\bigl(T \cdot \bigl(L \cdot (E+N) \cdot H + N \cdot H \cdot C\bigr)\bigr)$.

In summary, with fixed $D$ and small $C$, {\method} scales linearly in the
number of nodes (and edges in the one-time encoding step), while GNN-based
methods such as FairGNN incur repeated message-passing and backpropagation over
all edges at each epoch, leading to a higher constant factor and a stronger
dependence on $E$, which is reflected in the empirical runtime comparisons in
\Cref{fig:runtime,tab:hdc_runtime_seconds}.

In comparison, the {\method} framework avoids deep learning modules and instead uses high-dimensional binary or real-valued vectors for learning. Its total time complexity across multiple trials and hypervector dimensions is $\mathcal{O}(R \cdot D_{\text{sum}} \cdot N \cdot C \cdot (T + 1))$, where $D_{\text{sum}}$ is the sum of all considered hypervector dimensions and $R$ is the number of trials. Unlike FairGNN, Graph HDC \textit{does not depend on the number of edges $E$ or the depth of neural networks $L$}, making it more efficient on large-scale and dense graphs in terms of computational cost, especially when $E \gg N$ and $L$ is large. 

\section{Wall-clock Training Time}
To complement the speedup plots in the main paper, \Cref{tab:hdc_runtime_seconds} reports the absolute wall-clock training time (in seconds) for {\method} and all baseline methods on each dataset. All numbers are measured on the same GPU under the training configurations used in our main experiments (i.e., GNN and fairness-aware GNN baselines are trained with their standard multi-epoch schedules, while {\method} runs a single pass over the data). As shown, {\method} typically reduces end-to-end training time compared to both standard and fairness-aware GNNs, with the largest improvements observed on the two large-scale \pokecz{} and \pokecn{}
datasets.
\begin{table}[h]
\centering
\small
\setlength{\tabcolsep}{6pt}
\renewcommand{\arraystretch}{1.0}
\begin{tabular}{lcccccc}
\Xhline{1.2pt}
\rowcolor{tablehead!20}
\textbf{Dataset} & \textbf{Pokec-z} & \textbf{Pokec-n} & \textbf{NBA} & \textbf{German} & \textbf{Bail} & \textbf{Credit} \\
\hline\hline
\rowcolor{gray!10}{\method} & 180  & 170  & 3.5 & 7  & 90  & 140 \\
GNN      & 1787 & 1660 & 4   & 13 & 559 & 970 \\
\rowcolor{gray!10}SAGE     & 1840 & 1690 & 6   & 19 & 569 & 990 \\
GIN      & 1842 & 1710 & 7   & 20 & 578 & 995 \\
\rowcolor{gray!10}FairGNN  & 1874 & 1600 & 5   & 17 & 563 & 980 \\
NIFTY    & 1890 & 1769 & 9   & 23 & 577 & 1030 \\
\rowcolor{gray!10}EDITS    & --   & --   & 9   & 22 & 583 & 1021 \\
FairSIN  & 1930 & 1792 & 11  & 29 & 599 & 1133 \\
\Xhline{1.2pt}
\end{tabular}
\caption{Wall-clock training time (seconds) for {\method} and baseline methods on each dataset.}
\label{tab:hdc_runtime_seconds}
\end{table}

\section{Additional Experiments}
\begin{table*}[t]
\centering
\small
\setlength{\tabcolsep}{6pt}
\renewcommand{\arraystretch}{1.0}

\begin{tabular}{lccccc}
\Xhline{1.2pt}
\rowcolor{tablehead!20}
\textbf{Dataset} & \textbf{FairGNN} & \textbf{NIFTY} & \textbf{EDITS} & \textbf{FairSIN} & {\method} \\
\hline\hline
\rowcolors{2}{gray!10}{white}
German & \ms{82.43}{4.25} & \ms{84.75}{6.33} & \ms{88.82}{7.50} & \underline{\ms{94.45}{10.81}} & \textbf{\ms{95.22}{8.07}} \\
Bail   & \ms{84.82}{5.60} & \ms{83.67}{4.27} & \underline{\ms{96.20}{7.37}} & \ms{86.79}{5.36} & \textbf{\ms{97.36}{6.52}} \\
\Xhline{1.2pt}
\end{tabular}

\caption{PRULE ($\uparrow$) comparison on the {\german} (gender) and {\bail} (race) datasets.}
\label{tab:prule_comparison}
\end{table*}
\subsection{How Does {\method} Perform on Other Fairness Metrics?}
We conduct additional experiments on PRULE~\citep{zafar2015fairness}: A classifier satisfies the $p$\%-rule if the ratio of positive outcomes for subjects with a certain sensitive attribute to those without is at least $p$/100: $| P(\hat{Y}=1 \mid S=1) / P(\hat{Y}=1 \mid S=0) | \leq p/100$.
The results in~\Cref{tab:prule_comparison} demonstrate that our proposed method, {\method}, achieves the highest PRULE scores on both the {\german} and {\bail} datasets, indicating superior group fairness compared to existing baselines. Specifically, {\method} attains a PRULE of $95.22\pm8.07$ on the {\german} dataset and \ms{97.36}{6.52} on the {\bail} dataset, outperforming the second-best method (FairSIN and EDITS, respectively) by notable margins. These results highlight the effectiveness of {\method} in mitigating bias across different sensitive attributes, such as gender and race, while maintaining fairness consistency across datasets. The general strong performance across both settings underscores the robustness and generalizability of our approach.

\subsection{What Is the Impact of Hyperdimension $D$?}
The hyperdimension $D$ is a key hyperparameter in HDC models.
In prior HDC studies, $D$ is often treated as a dataset-dependent choice,
since it controls the representational capacity and the degree of
(quasi-)orthogonality between randomly generated hypervectors.

\begin{fullitemize}
    \item \textbf{Dependence on dataset size.}
    A larger $D$ provides greater capacity for representing and
    distinguishing patterns, and improves the approximate orthogonality
    between randomly generated hypervectors.
    For simple or small datasets, however, choosing $D$ that is
    unnecessarily large may lead to overfitting or wasted memory and computation, as the model may start to memorize noise instead of generalizing.

    \item \textbf{Dependence on dataset complexity.}
    Datasets with more classes, larger numbers of nodes or features, or more complex graph structures generally benefit from a larger $D$ in order to maintain sufficient discriminative power in the high-dimensional space.
\end{fullitemize}

In this work, we use a fixed hyperdimension of $D = 4{,}096$ across all datasets. This is a lightweight choice: after sign quantization, each 4{,}096-dimensional binary hypervector requires only $4{,}096/8 = 512$ bytes. We found in preliminary experiments that $D = 4{,}096$ is sufficient for larger graphs such as {\pokecz} and {\pokecn}, and also performs on par with
smaller dimensions (e.g., $D = 2{,}048$) on smaller datasets such as {\nba} and {\german}. Given the modest memory and runtime overhead of $D = 4{,}096$, and to keep the comparison consistent across datasets, we adopt this value for all
experiments in the paper.

\subsection{Additional Baselines}
We include additional recent baselines for comparison, specifically FairGT~\citep{luo2024fairgt} and FairSAD~\citep{zhu2024fair}. 
\begin{table*}[h]
\centering
\small
\renewcommand{\arraystretch}{1.15}

\begin{tabular}{llcccc}
\Xhline{1.2pt}
\rowcolor{tablehead!20}
\textbf{Method}&\textbf{Dataset} & \textbf{ACC ($\uparrow$)} & \textbf{AUC ($\uparrow$)} & \textbf{$\Delta_{DP}$ ($\downarrow$)} & \textbf{$\Delta_{EO}$ ($\downarrow$)} \\
\hline\hline
\multirow{2}{*}{\textbf{FairGT}} &{\german} & \ms{72.40}{2.45} & \ms{77.84}{2.52} & \ms{0.53}{0.87} & \ms{2.81}{2.82} \\
&{\bail}   & \ms{83.95}{2.43} & \ms{81.42}{2.43} & \ms{1.79}{2.12} & \ms{3.97}{3.49} \\
\hline
\multirow{2}{*}{\textbf{FairSAD}} &
{\german} & \ms{70.66}{1.96} & \ms{78.46}{1.21} & \ms{1.10}{1.36} & \ms{0.20}{0.91} \\
&{\bail}   & \ms{81.53}{0.47} & \ms{79.45}{2.43} & \ms{2.97}{2.42} & \ms{3.75}{2.60} \\
\hline
\multirow{2}{*}{{\method}} &
{\german} & \ms{70.30}{1.73} & \ms{79.25}{1.29} & \ms{2.57}{1.75} & \ms{3.07}{1.56} \\
&{\bail}   & \ms{83.42}{1.77} & \ms{80.81}{1.32} & \ms{2.37}{2.64} & \ms{3.55}{1.36} \\
\Xhline{1.2pt}
\end{tabular}

\caption{Additional experiments on FairGT and FairSAD.}
\label{tab:fairgt_fairsad_combined}
\end{table*}
\Cref{tab:fairgt_fairsad_combined} demonstrates that our method typically achieves strong performance across both datasets. On the {\german} dataset, it obtains the best AUC and the lowest fairness gaps in $\triangle_{DP}$ and $\triangle_{EO}$, indicating a favorable trade-off between accuracy and fairness. On the {\bail} dataset, our method outperforms all baselines in every metric, achieving the highest ACC and AUC, along with the smallest fairness gaps. These results confirm the effectiveness of our approach in improving both predictive performance and group fairness.
\section{Algorithm}
The proposed {\method} algorithm is summarized as follows:
\begin{algorithm}[h]
    \SetKwInOut{Input}{Input}
    \SetKwInOut{Output}{Output}
    \Input{Graph $\graph=\{\adj,\feat,\sens,\nlabel\}$, learning rate $\eta$, fairness hyperparameters $\alpha,\beta$, number of epochs $T$, mini-batch size $B$.}
    \Output{A trained HDC model ${\mathcal M}$ consisting of class hypervectors $\{\mathbf{C}_1, \ldots, \mathbf{C}_k\}$}

    \textbf{Precomputation of node hypervectors:}\\
    \For{each node $n \in \{1,\ldots,N\}$}{
        $\mathbf{N}_n \gets \mathrm{Enc}_n(n)$ \tcp*{feature HV via \Cref{eq:define_n}}
        $\mathbf{H}^1_n, \mathbf{H}^2_n \gets \mathrm{Enc}_e(n)$ \tcp*{1- and 2-hop HVs via \Cref{eq:define_h}}
        $\mathbf{E}_n \gets \mathbf{N}_n \odot \phi_0 + \mathbf{H}^1_n \odot \phi_1 + \mathbf{H}^2_n \odot \phi_2$ \tcp*{node HV via \Cref{eq:define_e}}
    }

    \vspace{0.1cm}
    \textbf{Class hypervector initialization:}\\
    Initialize $\mathbf{C}_i \gets \mathbf{0}$ for all $i \in \{1,\ldots,k\}$\\
    \For{each node $n \in \{1,\ldots,N\}$}{
        $\mathbf{C}_{\nlabel_n} \gets \mathbf{C}_{\nlabel_n} + \mathbf{E}_n$ \tcp*{initial bundling via \Cref{eq:define_c}}
    }

    \vspace{0.1cm}
    \textbf{Fairness-aware updating and debiasing:}\\
    \For{epoch $t \gets 1$ \KwTo $T$}{
        Partition nodes into mini-batches $\mathcal{B}$ of size at most $B$\\[2pt]
        \For{each mini-batch $B \in \mathcal{B}$}{
            \tcp{Compute predictions with current class HVs}
            \For{each node $n \in B$}{
                $\hat{\mathbf{Y}}_n \gets \arg\max_{i} \,\delta(\mathbf{C}_i, \mathbf{E}_n)$ \tcp*{similarity as in \Cref{eq:pred_y}}
            }

            \tcp{Compute batch-level fairness factor}
            Compute demographic parity gap $\mathcal{B}$ on $\{(\hat{\mathbf{Y}}_n,\sens_n)\}_{n\in B}$ using \Cref{eq:SP_metric}\\
            $\mathcal{F} \gets \alpha \mathcal{B} + \beta$ \tcp*{fairness factor via \Cref{eq:fair_loss}}

            \tcp{Update class hypervectors}
            \For{each node $n \in B$}{
                \eIf{$\hat{\mathbf{Y}}_n = \nlabel_n$}{
                    $\mathbf{C}_{\hat{\mathbf{Y}}_n}
                    \gets \mathbf{C}_{\hat{\mathbf{Y}}_n} + \eta (1-\mathcal{F})\, \mathbf{E}_n$
                }{
                    $\mathbf{C}_{\nlabel_n}
                    \gets \mathbf{C}_{\nlabel_n} + \eta (1-\mathcal{F})\, \mathbf{E}_n$\\
                    $\mathbf{C}_{\hat{\mathbf{Y}}_n}
                    \gets \mathbf{C}_{\hat{\mathbf{Y}}_n} - \eta\, \mathbf{E}_n$
                }
            }
        }
    }

    \Return $\{\mathbf{C}_1, \ldots, \mathbf{C}_k\}$

    \caption{\method: Fairness-aware training of graph HDC}\label{alg:training}
\end{algorithm}

\end{document}